# Integrating HMM-Based Speech Recognition With Direct Manipulation In A Multimodal Korean Natural Language Interface[*]

Geunbae Lee, Jong-Hyeok Lee, Sangeok Kim

Department of Computer Science & Engineering
Pohang University of Science & Technology
San 31, Hoja-Dong, Pohang, 790-784, Korea
Tel: +82-562-279-2254, Fax: +82-562-279-2299
`gblee@postech.ac.kr, jhlee@postech.ac.kr`

## ABSTRACT

This paper presents a HMM-based speech recognition engine and its integration into direct manipulation interfaces for Korean document editor. Speech recognition can reduce typical tedious and repetitive actions which are inevitable in standard GUIs (graphic user interfaces). Our system consists of general speech recognition engine called ABrain[1] and speech commandable document editor called SHE[2]. ABrain is a phoneme-based speech recognition engine which shows up to 97% of discrete command recognition rate. SHE is a EuroBridge widget-based document editor that supports speech commands as well as direct manipulation interfaces.

**Keywords** - speech recognition, direct manipulation, multimodal interface, document editor

## INTRODUCTION

Recent development in graphical user interface (GUI) provides pull-down menu, direct manipulation [8], and icons for more user-friendly human interactions. However, direct manipulation (DM) style GUIs still have many points to be improved:

- Current DM interfaces put so much cognitive load to users to find the targets among the huge number of possible items. Moreover, due to the limited screen size, all the necessary items cannot be displayed for the user selection.

- The DM interfaces require repetitive user actions for series of similar interactions. Compared with command interfaces, they lack macro functions and flexibilities through parameter specification.

- The DM interfaces require very accurate motor control at users' side, so they incur frequent muscle fatigues. Moreover, they are simply inadequate for most of the disabled persons.

Many of the problems of DM interfaces can be resolved using the speech inputs:

- Users don't have to find the targets among the huge number of items. They just remember the command and speak it, and the search is automatically performed by the computer.

- Users don't repeat the same actions. They just select the target and speak the command with the number options.

- Users can avoid accurate motor control fatigue since they put the pointer around the target and the correct targets can be selected according to the speech command.

Today, speech recognition technology matures enough to build under 500 word command interpreter that can be efficiently used in any GUI environment. However, speech interface alone also has the following problems:

- Poorly designed speech interface has usability problems. For example, the cursor movements are very awkward for speech commands.

- The speech interface has the same problems as the command interface, that is, users have to memorize the commands.

- The response time usually becomes slower than keyboard or mouse.

- The speech command can be mis-recognized so proper confirmation process is always required.

Because of the current limitations of speech technology, usabilities of the speech interface should be improved by integrating it with other input modalities such as DM interfaces. In this paper, we aim to develop HMM-based speech recognizer and integrate it into the DM interfaces for Korean document editor. To overcome the problems of speech only interfaces or standard GUIs including the DM, we decided to split tasks between the two input modalities: The object selection is performed using the DM interfaces, while the action specification is performed using the speech. This split of tasks follows the suggestion that the spatial task can be best served by the visual input and manual responses while the verbal tasks can be by the auditory input and speech responses [10]. With the study of $MacDraw^{TM}$, the voice command integrated with DM was reported to reduce the actual task completion times [4]. Our integrated speech and DM interfaces are developed to be used in Korean document editing environment. For this task, a prototype Korean editor was also developed that can support multiple Korean fonts.

[*]This research was supported by PIRL (Pohang Information Research Laboratory)
[1]Auditory Brain
[2]Simple Hearing Editor

## SYSTEM ARCHITECTURE

The system consists of two parts: ABrain, a speech recognizer and SHE, a Korean document editor which supports both speech and DM (direct manipulation). Figure 1 shows the overall architecture of the system. SHE communicates with both ABrain and X-window server to support speech and DM interfaces simultaneously.

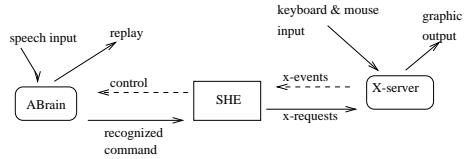

**Figure 1:** System architecture of multimodal Korean document editor

## HMM-Based Speech Recognition

The speech recognizer ABrain is a HMM-based speech recognition engine which employs phonemes as basic recognition units. It is a speaker-dependent discretes-speech recognizer with medium-sized vocabulary. Figure 2 shows the overall flow of HMM speech recognition. The voice sampling is based on 16kHz-16bit standard encoding. We use explicit style end-point detection with STE (short time energy) and ZCR (zero crossing rate) analysis for typical isolated word HMM recognition. The detected word frames are pre-emphasized for high frequency compensation, and are hamming-windowed for smoothing. The feature vectors are generated based on the 15 mel-scaled filter-bank parameters. Figure 3 shows feature vector generation scheme. Here, n is a frame size (16 msec interval in 16kHz sampling), and 3 different feature vectors are generated for single frame. ABrain uses 3 different codebooks to reduce the quantization errors since it is reported that large feature vectors increase the quantization errors[2]. The 3 different codebooks are based on 3 different feature vectors in figure 3.

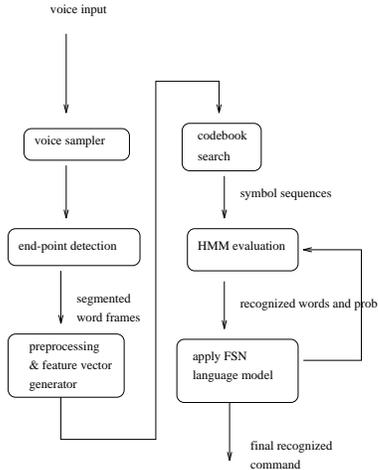

**Figure 2:** ABrain speech recognition process

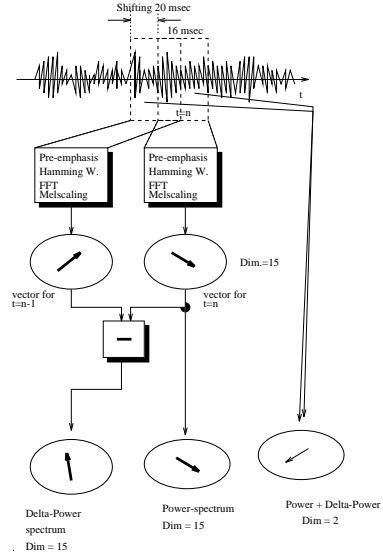

**Figure 3:** Feature vector generation from mel-scaled filter-bank energy

ABrain has a basic HMM model for each of the 39 Korean phonemes. The phoneme HMM model has very simple structure: it has 3 different states and only left-to-right and self transitions. Usually, HMM phoneme model training requires large amount of segmented speech data. However, there is no standard Korean speech database available yet. So we employed segmental k-means training [5] since the algorithm automatically segments the data while performing training. In this way, we can efficiently train our HMM models without large amount of segmented speech database. Figure 4 shows the training procedure. To initialize the phoneme HMM, we usually use small sized pre-segmented data for fast convergence. The level-building algorithm [3] automatically segments the words into phone boundaries. Using the phone boundary and state information, we collected all the same labeled codeword sequences and re-estimate the HMM parameters according to the following (using k-means clustering hence the name segmental k-means training):

$$a_{ij} = \frac{\#\ of\ transitions\ from\ state\ i\ to\ j}{\#\ of\ transitions\ from\ state\ i\ to\ any\ state} \quad (1)$$

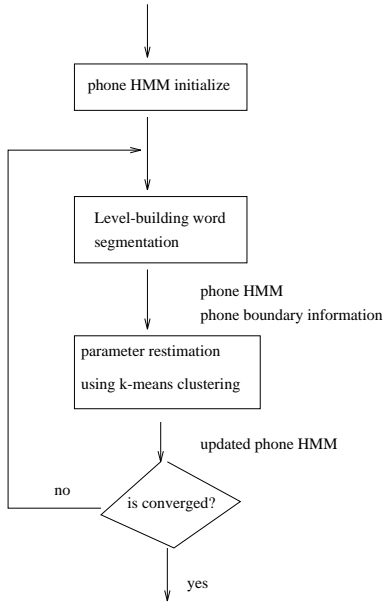

**Figure 4:** Segmental k-means training of the phoneme HMM model

$$b_i(k) = \frac{\#\ of\ vectors\ with\ codeword\ k\ in\ state\ i}{\#\ of\ vectors\ in\ state\ i} \qquad (2)$$

where $a_{ij}$ and $b_i(k)$ are updated HMM transition and emission probabilities respectively. The training continues until the updated HMM models converge. At the recognition phase, ABrain uses phonetic dictionary to build word HMM model and performs frame synchronous Viterbi search to find the proper word sequence for the command. The explicit silence HMM model is used between the word. ABrain also employs finite state language model for a sequence of document editor commands. The number of possible words are about 200, and the possible command sequences are over 300 in our system. The word-pair language model is applied through the Viterbi search.

## WYSIWYG Korean Editor

We developed WYSIWYG (What You See Is What You Get)-style Korean document editor based on EuroBridge Xew widget[3]. Xew widget supports multimedia such as text, image, voice and video. Among these, the text widget conforms to ISO 2022 code extension, so we can display Korean character (hangul). However, for hangul input, we implemented hangul automata. Figure 5 shows the snapshot of hangul display in Korean document editor SHE (simple hearing editor).

SHE communicates to ABrain speech recognizer through shared memory and semaphore mechanism in UNIX. Fig-ure 6 shows the communication between SHE and ABrain. When ABrain is activated by calling its name (e.g. "kant"), it starts handshaking with SHE and gets into the sleep state until the semaphore flag is unset. SHE notifies the user that ABrain is activated and unset the flag. Now user can issue the voice command (e.g. "kulccasayk" meaning "font color") to ABrain, and SHE performs the action and sets the flag so that ABrain gets into the sleep state until the user wakes it up by calling its name again. In figure 6, the number shows the execution flow of the handshaking.

---

[3]EuroBridge widget is a multimedia application toolkit based on the X-window and Xt library, and developed during the RACE-2008 project at VTT (Valtion Teknillinen Tutkimuskeskus) in Finland.

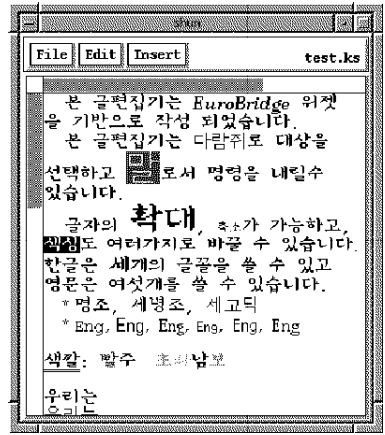

**Figure 5:** Korean document editor SHE based on X-window

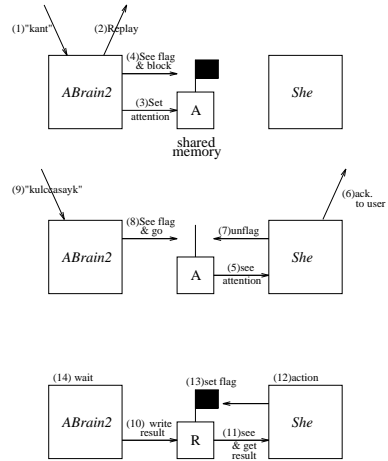

**Figure 6:** Communication between ABrain and SHE

## EXPERIMENTS

ABrain's 39 phoneme models are trained using 687 words, and tested for isolated speaker-dependent performance for new 198 words without any language model (perplexity 198). In this case, ABrain shows medium performance of 55% recognition rate. The top 3 and top 6 performances are 78% and 90% respectively. We built a finite state language model for SHE commands using these 198 words. For over 300 commands, under the speaker-dependent isolated word condition, we obtained 97% recognition rate and 2 sec response time which are reasonable performances for editing task. We are now experimenting on the usabilities of multimodal editing using both speech and direct manipulation regarding the ABrain and SHE.

| item | ours | NEC | XspeakII |
|------|------|-----|----------|
| DM | yes | yes | yes |
| win nav. | no | no | yes |
| ol tra. | no | don't know | yes |
| ded. sp. | no | yes | yes |
| LM | FSN | don't know | similar to FSN |

**Table 1:** Comparison of the 3 systems in each of the items: DM (direct manipulation), win nav. (window navigation), ol tra. (on-line training), ded. sp. (dedicated speech server), and LM (language model). Xspeak II uses contexts-dependent word recognition but the concept is similar to our FSN language model.

## COMPARISON WITH RELATED RESEARCHES

There have been several researches on multimodal interfaces. As for the voice, [6] studied the variables of a user performing a task using voice by developing voice spreadsheet and observing users interacting with it. As for the integration of natural language and general pointing, [9] analyzed the effects of general pointing to the discourse models of multimodal interaction. To directly compare our system with others, we will discuss two previous systems that are most similar to ours in integrating speech and direct manipulation. The NEC multimodal drawing tool [1] integrates keyboard, mouse, and speech inputs in a drawing tool and presents a new method of interpretation of pointing with voice depending on the context. It is similar to ours that it can allow users to use both speech and DM freely at user's disposal. But our system uses shared memory and semaphore for communication between speech recognizer and document editor while the NEC drawing tool has separate input integrator to connect to the drawing tool. The Xspeak II [7] system allows users to do window navigation using voice commands. The Xspeak team develops specific language X-GL to generate several X-events including voice events, and the voice events are treated samely as the other normal x-events. Xspeak II is similar to our system that it is based on X-window, but our system can communicate with any general speech recognizer rather than with a dedicated speech server. Table 1 shows the comparison between our system with the two above mentioned systems.

## CONCLUSION AND FUTURE WORKS

In this paper, we developed primitive multimodal Korean document editor supporting both speech and DM interfaces. For efficient synergistic integration, we split user's tasks into two parts: selection and action, and used DM for selection and speech for action specification. The speech recognizer ABrain is newly developed for Korean using phoneme-level HMM model and segmental k-means training. The Korean document editor SHE was developed based on Xew widget and extended to accept speech inputs as well as mouse/keyboard inputs. The communication between ABrain and SHE was done using the UNIX shared memory and semaphore. However, ABrain speech recognizer has many rooms to improve. The phoneme model should be improved to handle the phonemes with short frames (less than 3 frames). Also speaker independence, continuous speech, robust with noise, and unrestricted natural language handling are essential features of any future speech recognition systems. SHE should be more experimentally tested with human subjects to prove the usabilities of multimodality in document editing tasks, especially of the effects of speech with DM environments.